\documentclass[12pt,a4paper]{article}
\usepackage[utf8]{inputenc}
\usepackage{amsmath}
\usepackage{amsfonts}
\usepackage{amssymb}
\usepackage{graphicx}
\usepackage{url}
\usepackage{morefloats}
\usepackage{multirow}
\usepackage[table,xcdraw]{xcolor}
\usepackage[left=2cm,right=2cm,top=2cm,bottom=2cm]{geometry}
\begin{document}
\textbf{An Interactive Medical Image Segmentation Framework Using Iterative Refinement}\\
Pratik Kalshetti, Manas Bundele, Parag Rahangdale, Dinesh Jangra, Chiranjoy Chattopadhyay, Gaurav Harit, Abhay Elhence
\begin{abstract}
Image segmentation is often performed on medical images for identifying diseases in clinical evaluation. Hence it has become one of the major research areas. Conventional image segmentation techniques are unable to provide satisfactory segmentation results for medical images as they contain irregularities. They need to be pre-processed before segmentation. In order to obtain the most suitable method for medical image segmentation, we propose a two stage algorithm. The first stage automatically generates a binary marker image of the region of interest using mathematical morphology. This marker serves as the mask image for the second stage which uses GrabCut on the input image thus resulting in an efficient segmented result. The obtained result can be further refined by user interaction which can be done using the Graphical User Interface (GUI). Experimental results show that the proposed method is accurate and provides satisfactory segmentation results with minimum user interaction on medical as well as natural images.
\end{abstract}
\section{Introduction}
Magnetic Resonance Imaging (MRI) and X-Ray Computed Tomography (X-ray CT) yields a series of images generating a volume data which are viewed by medical professional for diagnosis, treatment planning or population studies \cite{ref3}. Owing to various restrictions imposed by image acquisitions, pathology and biological variation \cite{ref1}, the medical images are of high complexity and ambiguity as well as rich in noise and low in contrast \cite{ref2}. This makes segmentation of the region of interest from these images a difficult process. Segmentation is the task of identifying and localizing salient structures in the image volume \cite{ref3}. Traditionally this process is done manually slice by slice, which requires expert knowledge to obtain a particular anatomic region or organ of interest from the volume data \cite{ref1}. Due to infeasible time and cost of manual segmentation, a number of computer-aided segmentation techniques are being developed for medical images \cite{ref1}. Large amounts of data coupled with time and cost constraints have led to the development of automated segmentation techniques and a part of recent medical computing literature \cite{ref3}. The automated methods of segmentation provide results without prior knowledge about the images and do not require human interaction \cite{ref1}. However a human expert's ability to combine observed image data with his prior knowledge is unmatched when compared with the computer algorithms. So, considering this fact, interactive segmentation methods are a recent field of study in the medical image analysis \cite{ref3}. In our paper, we refer to Magnetic Resonance (MR) and X-Ray images as medical images.\par
So far, many segmentation techniques have been studied by the researchers which can be classified based on threshold, edge, fuzzy theory, partial differential equations (PDE), artificial neural networks (ANN), region \cite{ref5} and graph \cite{ref2}. Threshold based methods \cite{ref6} do not work well for images without obvious peaks or with broad and flat valleys whereas the edge based techniques work well for images having contrast between regions but does not work well with images having high noise and ill-defined edges. The Fuzzy theory based techniques like Fuzzy C means \cite{ref4} have huge computation time and do not produce standard segmentation result always due to the random nature of initial membership values. The PDE based techniques \cite{ref4} such as deformable models, when applied to noisy images with ill-defined boundary, may produce shapes that have inconsistent topology with respect to the actual object. Although the ANN based methods \cite{ref4} possess capability of self-organization due to the learning through training data and can process in real time due to parallel configuration, they have a few drawbacks. The region based segmentation techniques are by nature sequential and quite expensive both in computational time and memory. Also, region growing \cite{ref6} has inherent dependence on the selection of seed region and the order in which the pixels and regions are examined. Watershed based segmentation \cite{ref4} groups pixels of an image on the basis of their intensities but it is sensitive to intensity variations and results in oversegmentation. Graph cuts \cite{ref7} is a graph based combinatorial optimization technique applied to the process of image segmentation by Boykov and Jolly in 2001. One of the advantages of graph based technique is that it might require no discretization by virtue of purely combinatorial operators and thus have no discretization errors \cite{ref8}. Graph cuts based methods \cite{ref7}treat image segmentation problem as a min-cut problem on graphs where each image pixel is treated as a graph node. The GrabCut algorithm \cite{ref7} is an extension of Graph cuts which works iteratively in the energy minimization process. After the analysis of various techniques, it has been observed that a hybrid solution consisting of two or more techniques for segmentation is the best approach for solving the problem of medical image segmentation \cite{ref5}.\par 
In this paper, we have proposed an interactive medical image segmentation technique which uses a combination of mathematical morphology and GrabCut algorithm. Mathematical morphology is used as a pre-processing technique whose output is given to the GrabCut algorithm, thus giving an efficient segmentation result. Also interaction is provided so as to incorporate application specific needs of the user.\par
\section{Contribution}
As mentioned before, traditionally segmentation was done manually by the experts which is very tedious, inaccurate and subjective from user to user. Fully automatic segmentation with high accuracy still remains an open problem in the field of medical image segmentation especially when the images consist of large amounts of noise \cite{ref10}. Thus, interactive segmentation techniques serve as a perfect platform where the doctors can edit the results as per their convenience. However, automatic techniques \cite{ref1} do provide segmentation results without prior knowledge about the images and can generate rough segmentation results and can thus be used for pre-processing of the images. The contribution of the proposed technique revolves around the subsequent sections.\par
\subsection{Automatic Initialization}
We have used mathematical morphology for automatic initialization which generates a marker image that can further serve as an input to the second stage which uses GrabCut. The GrabCut algorithm takes input in the form of seed points through marker image or a rectangle drawn around the object of interest \cite{ref7}. Manually generating marker image for each and every image slice is still time consuming and thus can be automated to speed up the process. Also the marker generated consists of stray isolated pixels and some shadowed objects are not marked. The obtained marker is modified by erosion and removing components having less than a particular number of pixels. The results of the generated marker image using morphology are shown in 
\begin{figure}[t]
\centering
\includegraphics[scale=0.2]{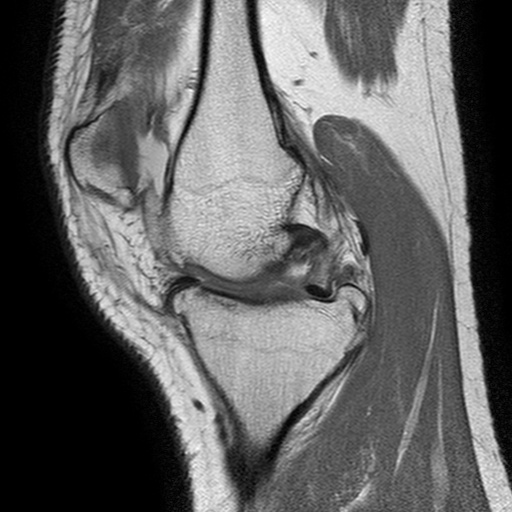}
\includegraphics[scale=0.2]{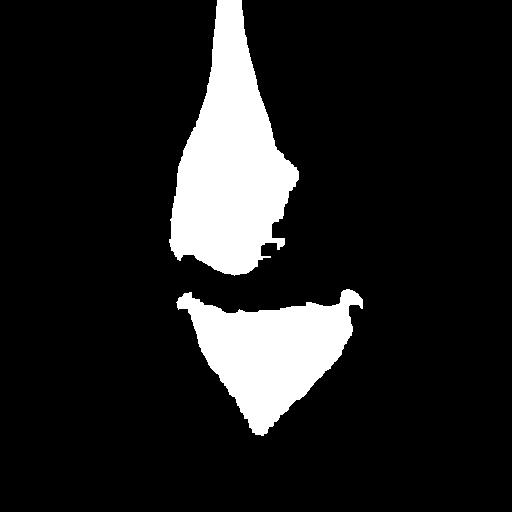}
\caption{Modified marker (right) of knee MR image (left) obtained by automatic initialization. }
\end{figure}

\subsection{High Accuracy}
A hybrid solution of using morphology as a pre-processing technique for GrabCut algorithm provides more accurate results when compared with results obtained using only GrabCut. The marker image generated using morphological operations rule out the possibility of unnecessary foreground object’s detection. This marker image when given to GrabCut gives more accurate results as the foreground(using marker) and background(using bounding box) are initially marked properly as well as automatically. Thus, morphology driven automatic initialization helps in generating a more accurate result.
\begin{figure}[H]
\centering
\includegraphics[scale=0.26]{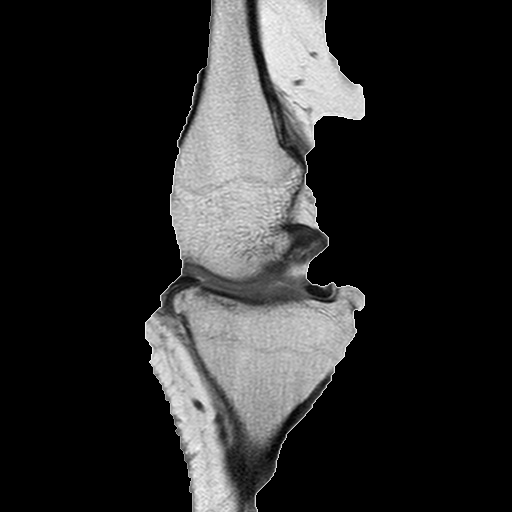}
\includegraphics[scale=0.195]{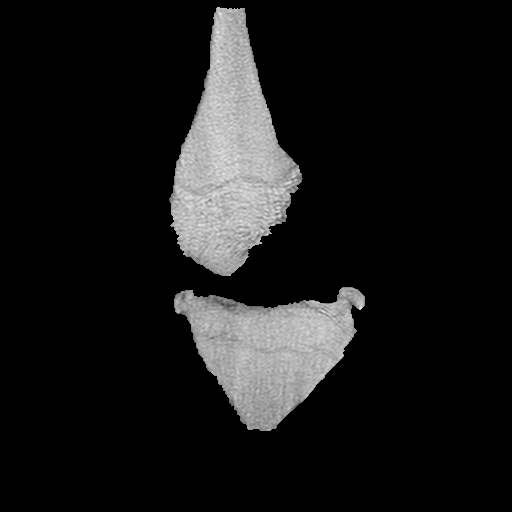}
\caption{Comparison of segmented results for GrabCut (left) and MIST (right).}
\end{figure}

\subsection{Minimal Interaction}
Since the results generated from automatic initialization and GrabCut output are way more accurate as compare to only GrabCut results, there is a much less need for any kind of interaction to get refined results. Thus, doctors have to spend less time editing the obtained results and thus more time can be used for the analysis of the patient data.
\begin{figure}[H]
\centering
\includegraphics[scale=0.26]{grabcutcompare.png}
\includegraphics[scale=0.195]{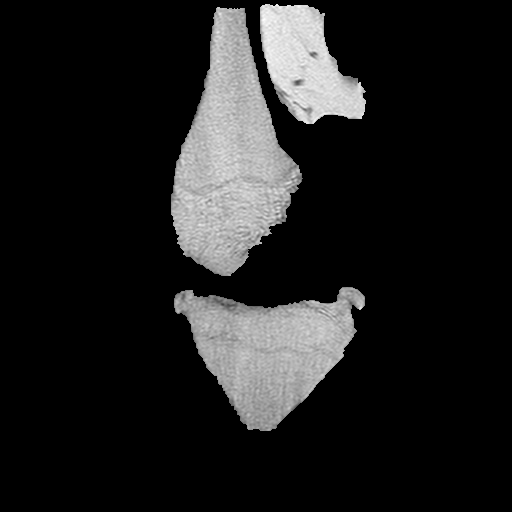}
\caption{Comparison of results: GrabCut (left) and MIST before user interaction (right).}
\end{figure}

\subsection{Generalized for natural images}
The proposed algorithm has not only proved to be accurate for the medical image data but also has worked well for natural images. In Fig. \ref{plant}, the results of MIST when applied to a natural image are given:
\begin{figure}[t]
\centering
\includegraphics[scale=0.2]{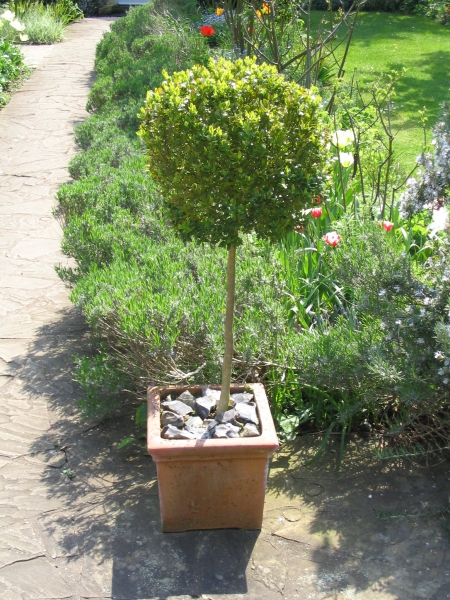}
\includegraphics[scale=0.2]{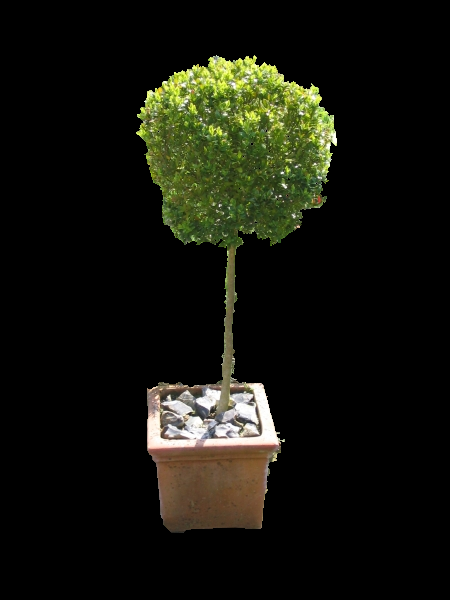}
\caption{Input image (left) and Segmented output (right).}
\label{plant}
\end{figure}

In Fig. \ref{plant}, we can see that even though the grass in the background has the color (green) similar to the leaves of the plant, the plant with its pot and stem are properly segmented out from the image with minor user refinements.\\\\ 
The rest of the paper is organized as follows: In Sec. \ref{section2}, a brief description of the framework is provided. Sec. \ref{section3} comprises of the results and discussion. Sec. \ref{section4} refers to the conclusion and Sec. \ref{section5} depicts the Graphical User Interface of MIST.

\section{Brief Description of the Proposed Framework} 
\label{section2}

\begin{figure}[t]
\centering
 \includegraphics[scale=0.6]{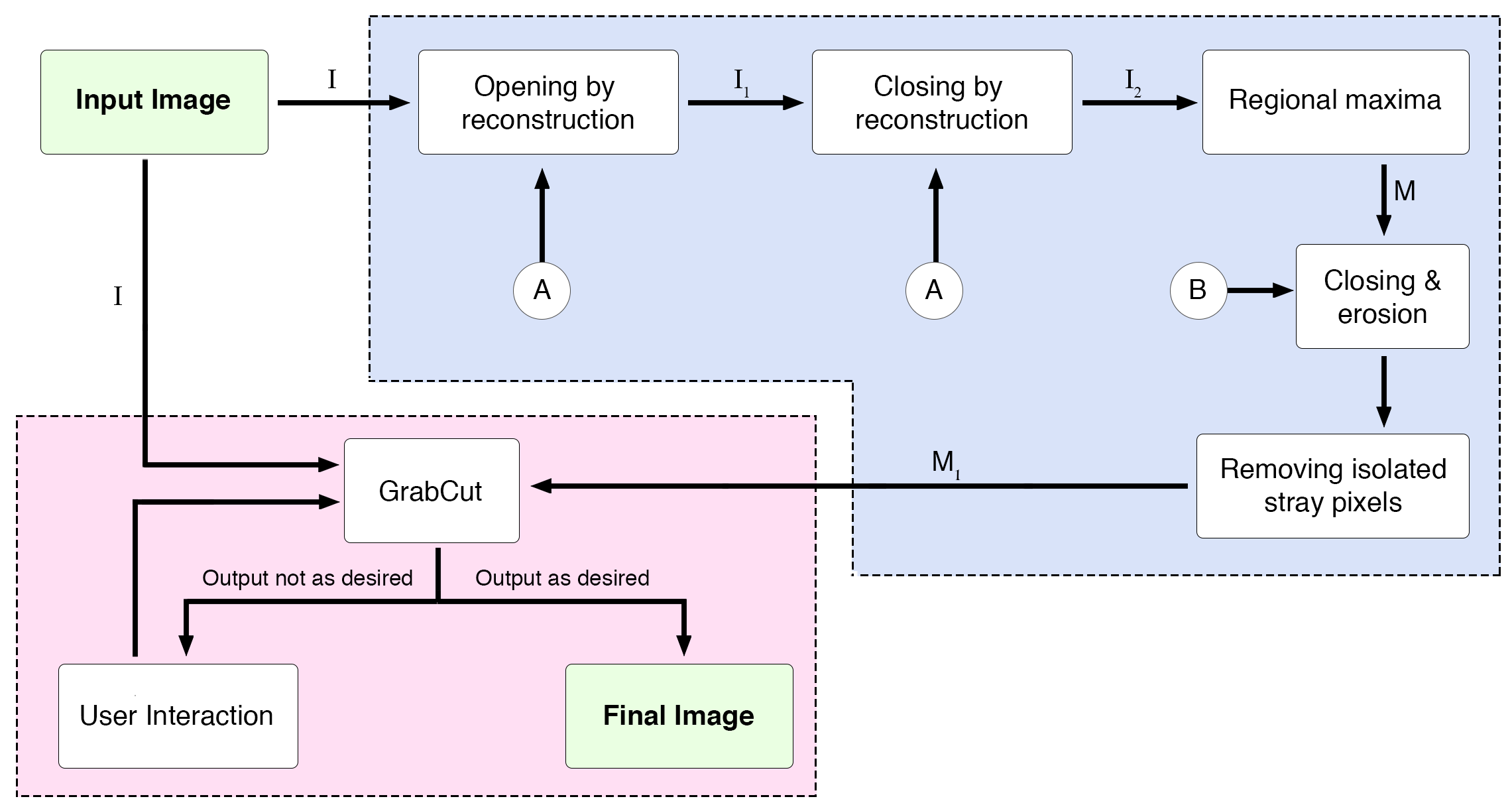}
 \caption{Framework of MIST.}

 \label{framework}
\end{figure}
\par

In Fig. \ref{framework}, the framework of MIST is shown in the form of steps involved.  The stage one of the algorithm highlighted by blue color represents the marker generation process while the one in pink represents stage two involving the GrabCut algorithm and the User Refinement. The input image and the output of the first stage are fed as inputs to the GrabCut algorithm which provides the final output based on the need for user refinement. The input image $I$ gives $I_1$ as the output on performing an opening by reconstruction using the structuring element $A$ where $I_1 = O_R^{1}(I)$ and $O$ in $O_R^{1}$ denotes opening, the subscript $R$ shows reconstruction and the superscript $1$ shows the number of erosions used in the process. The structuring element $A$ is disk shaped with variable radius. $I_1$ when operated by $C_R^{1}$ gives $I_2$ as output on performing closing by reconstruction by the same structuring element $A$ where $C$ in $C_R^{1}$ denotes closing, the subscript $R$ shows reconstruction and the superscript $1$ denotes the number of dilations in the process. The regional maxima of $I_2$ gives an initial marker image $M$ as output. This initial marker image $M$ is further acted upon by closing and erosion by structuring element $B$ ($5 \times 5$ matrix containing all ones) to correct the marking near the edges from which the stray isolated pixels are removed to obtain a modified marker image $M_1$. The structuring element $A$ is disk shaped with radius depending on the size of input image. This modified marker image $M_1$ and the input image $I$ is then given to the GrabCut algorithm which provides us the initial segmented output. Based on the desired output, the initial segmented output is then further subjected to user interaction for refinement. On the basis of the labelled pixels by the user, the GrabCut algorithm is re-run, thus giving the final output.

\subsection{Dataset Description}
\begin{figure}[t]
\centering
 \includegraphics[scale=0.2]{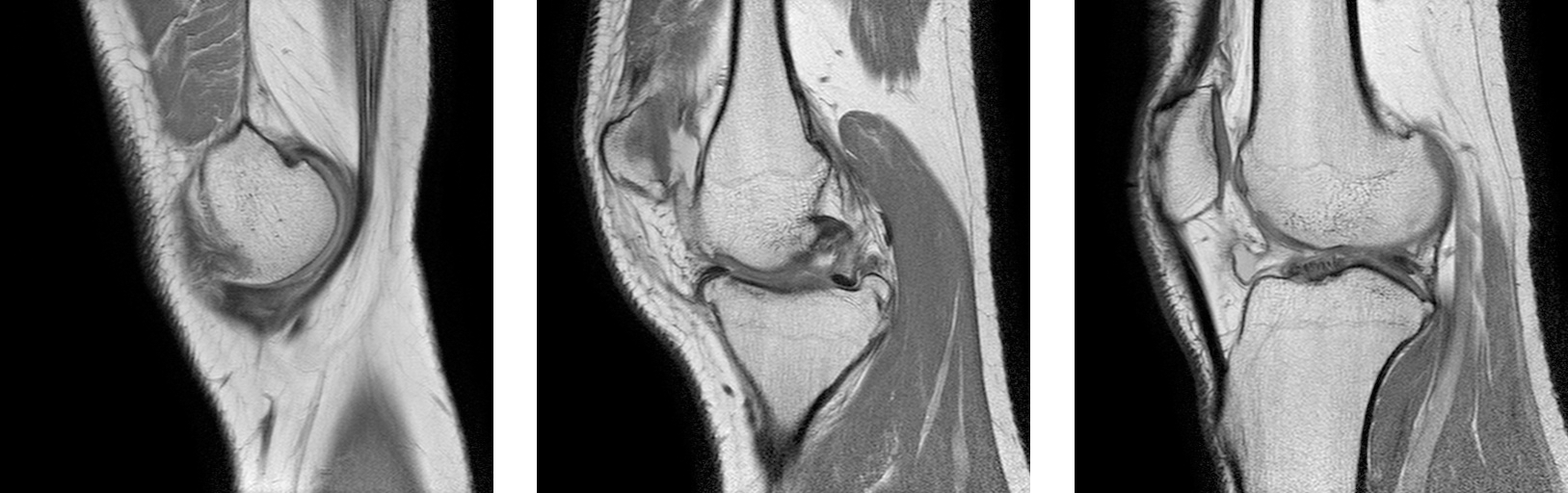}
 \caption{Three different input MR images of size $512 \times 512$.}
 \label{input}
 \end{figure}

The medical images used in the dataset consist of DICOM images. Digital Imaging and Communication in Medicine (DICOM)  \cite{ref15} is a standard for storing, handling and transmitting information in medical imaging. 
DICOM includes data structures for medical images and associated data, network oriented services for image transfer or printing, media formats for data exchange, work-flow management, consistency and quality of presentation. These DICOM images from our dataset are of size $512\times 512$ as shown in Fig. \ref{input}.

\par

\subsection{Automatic Marker Generation}
Raw medical image data is the information obtained during an experiment, before the information has been analysed or statistically manipulated \cite{rawdata}. It has high amount of noise. Due to this reason, image segmentation techniques don't work well on raw medical images. In MIST, the smoothing of these raw medical images is carried out using morphological reconstruction \cite{ref21}. The resultant image is then used to generate a marker that acts as a mask for the stage two of the framework. A disk shaped structuring element $A$ is chosen for the morphological reconstruction phase whose size is chosen as per the dataset under consideration.

\begin{figure}
\centering
  \includegraphics[width=.4\linewidth]{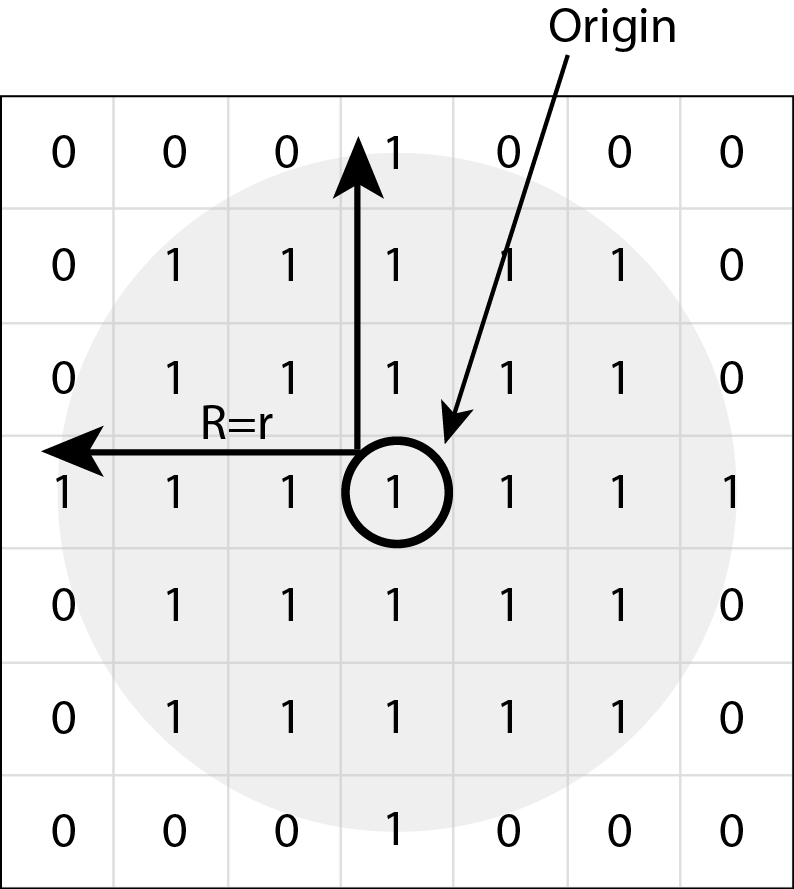}
  \includegraphics[width=.4\linewidth]{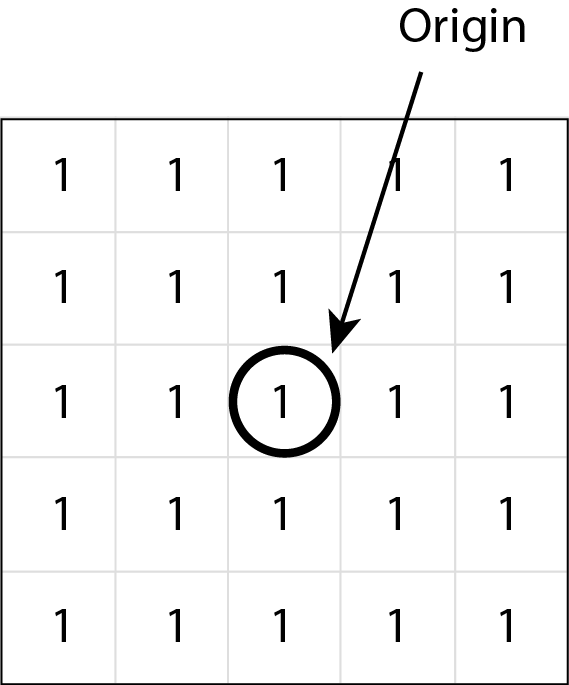}
  \caption{Structuring elements: $A$ (left) with variable $r$ and $B$ (right) with size $5 \times 5$.}
  \label{structelem}
\end{figure}

Because of the isotropic nature of the disc-shaped structural element, they are able to eliminate the dependency of gradient on the edge-orientation. The structural element has a small radius so that the excessive thickening of edges can be avoided.\par
In MIST, initially opening-by-reconstruction is operated on the the input image and then followed by closing-by-reconstruction. This eliminates the regional extrema caused by gray-scale image irregularities, thus reducing the amount of disturbance and noise present in the input image. Another advantage of using the morphological operation is that the extrema of important contours are preserved during the process. Traditional morphological opening and closing operation can only remove regional details in parts of high and low gray-scale pixels in images. On the other hand, opening closing by reconstruction operation in the process of smoothing images can completely remove or retain regional details smaller than the current size in high and low gray-scale regions.\cite{ref9}\par
\begin{flushleft}
In morphological opening-by-reconstruction, initially an erosion operation on the input image is done followed by reconstruction-by-dilation. The eroded image $I_e$ is given as:\\
\end{flushleft}
\begin{equation} \label{eq1}
%\begin{split}
I_e = I \ominus A
%\end{split}
\end{equation}

where $\ominus$ defines the erosion operator, $I$ denotes the input image and $A$ is the structuring element used as defined in Fig. \ref{structelem}.\\
Erosion is followed by the morphological reconstruction-by-dilation using the eroded image $I_e$ as a marker and the input image $I$ as a mask. The morphological reconstruction-by-dilation is the repeated dilation of a marker image until the contour of the marker image fits under the mask image. This step helps in removing regional extremity associated with regional minima. The reconstruction-by-dilation is given by:\par
\begin{equation} \label{eq2}
R_I^{D}(I_e) = D_I^{n}(I_e)
\end{equation}
where $D_I(I_e)$ is the geodesic dilation of $I_e$ (Eq. \ref{eq1}) with respect to $I$, assuming $n$ iterations are required to achieve stability, where $D_I^{n}(I_e)$ is defined by: 
\begin{equation} \label{eq3}
D_I^{n}(I_e) = D_I^{1}[D_I^{n-1}(I_e)]
\end{equation}
and 
\begin{equation} \label{eq4}
D_I^{1}(I_e) = (I_e \oplus A) \cap I
\end{equation}
where $\oplus$ is the dilation operator.\par

Finally the result after opening-by-reconstruction is given by:
\begin{equation} \label{eq5}
I_1 = O_R^{1}(I) = R_I^{D}(I_e)
\end{equation}

\begin{figure}[t]
\centering
\includegraphics[scale=0.15]{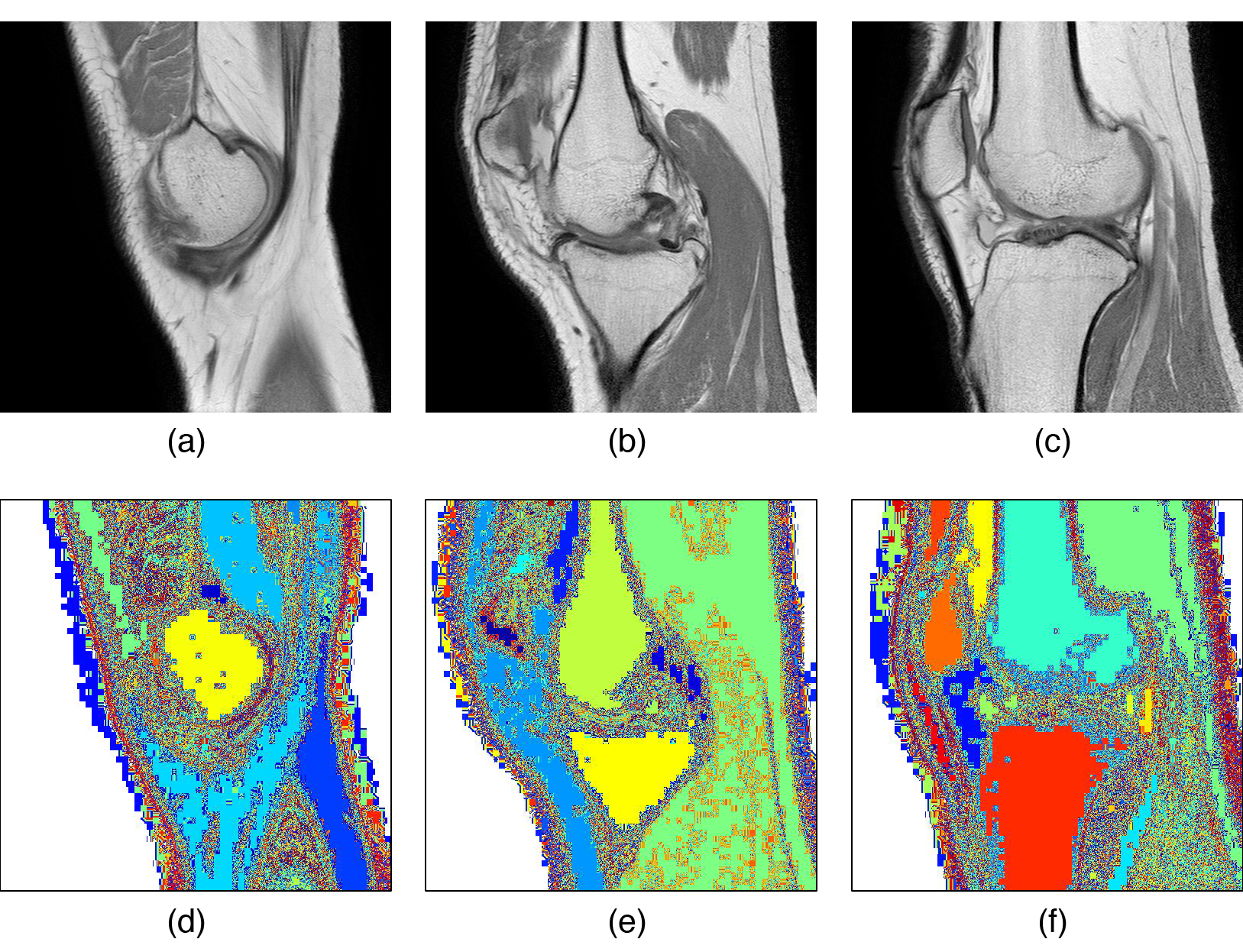}
 \caption{(a), (b) \& (c): Different input images; (d), (e) \& (f): Output of opening-by-reconstruction, i.e. $I_1$ (Eq. \ref{eq5}) on above respective input images.}
\label{open}
\end{figure}\par

In Fig. \ref{open}, pseudo-coloring is performed on the output gray-scale image of opening-by-reconstruction for better visualization. Since the gray-scale image is blurred after opening by reconstruction, it shows smaller regions of different colors on pseudo-coloring.
Opening-by-reconstruction is followed by morphological closing-by-reconstruction. In this method, we perform dilation operation on output image $I_1$ using the same structuring element $A$. The dilation on input image $I_1$ to give $I_d$ which is defined as:\\
\begin{equation} \label{eq6}
%\begin{split}
 I_d = I_1 \oplus A
 %\end{split}
 \end{equation}

Dilation is followed by morphological reconstruction-by-closing. We use the complement of input image given as $I_c$ as a marker and the complement of the dilated image mask given by $I_{dc}$. This step helps in removing regional extremity associated with regional minima. The reconstruction-by-closing is given by:\\
\begin{equation} \label{eq7}
R_{I_c}^{E}(I_{dc}) = E_{I_c}^{n}(I_{dc})
\end{equation}
where $E_{I_c}(I_{dc})$ is the geodesic erosion of $I_{dc}$ with respect to $I_c$, assuming $n$ iterations are required to achieve stability, where $E_{I_c}^{n}(I_{dc})$ is defined by: 
\begin{equation} \label{eq8}
E_{I_c}^{n}(I_{dc}) =E_{I_c}^{1}[E_{I_c}^{n-1}(I_{dc})]
\end{equation}
and 
\begin{equation} \label{eq9}
E_{I_c}^{1}(I_{dc}) = (I_{dc} \ominus A) \cup I_c
\end{equation}
where $\ominus$ is the erosion operator.\par

Finally the result after closing-by-reconstruction is given by:
\begin{equation} \label{eq10}
I_2 = C_R^{1}(I_1) = R_{I_c}^{E}(I_{dc})
\end{equation}

\begin{figure}[t]
\centering
 \includegraphics[scale=0.15]{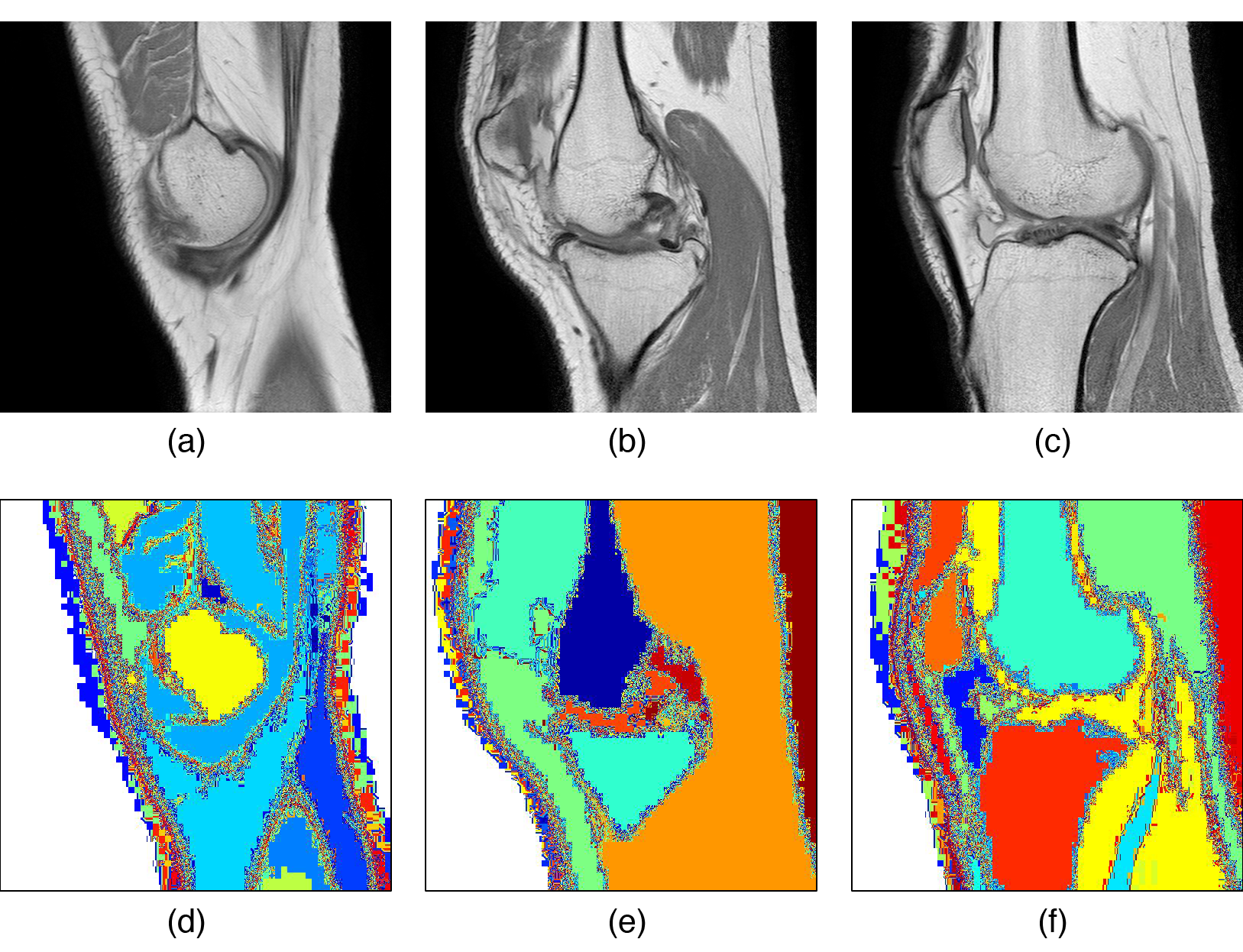}
 \caption{(a), (b) \& (c): Different input images; (d), (e) \& (f): Output of opening closing by reconstruction, i.e. $I_2$ (Eq. \ref{eq10}) on above respective input images.}
\label{openclose} 
 \end{figure}
\par
The gray-scale image is further smoothened on performing opening closing by reconstruction. Thus, Fig. \ref{openclose} is more smooth and has less pixel dissimilarities as compared to Fig. \ref{open}.\\ 
These operations will create a flat maxima, removing small blemishes without affecting the overall shapes of the objects. The regional maxima provides an initial marker $M$.
\begin{figure}[t]
\centering
 \includegraphics[scale=0.15]{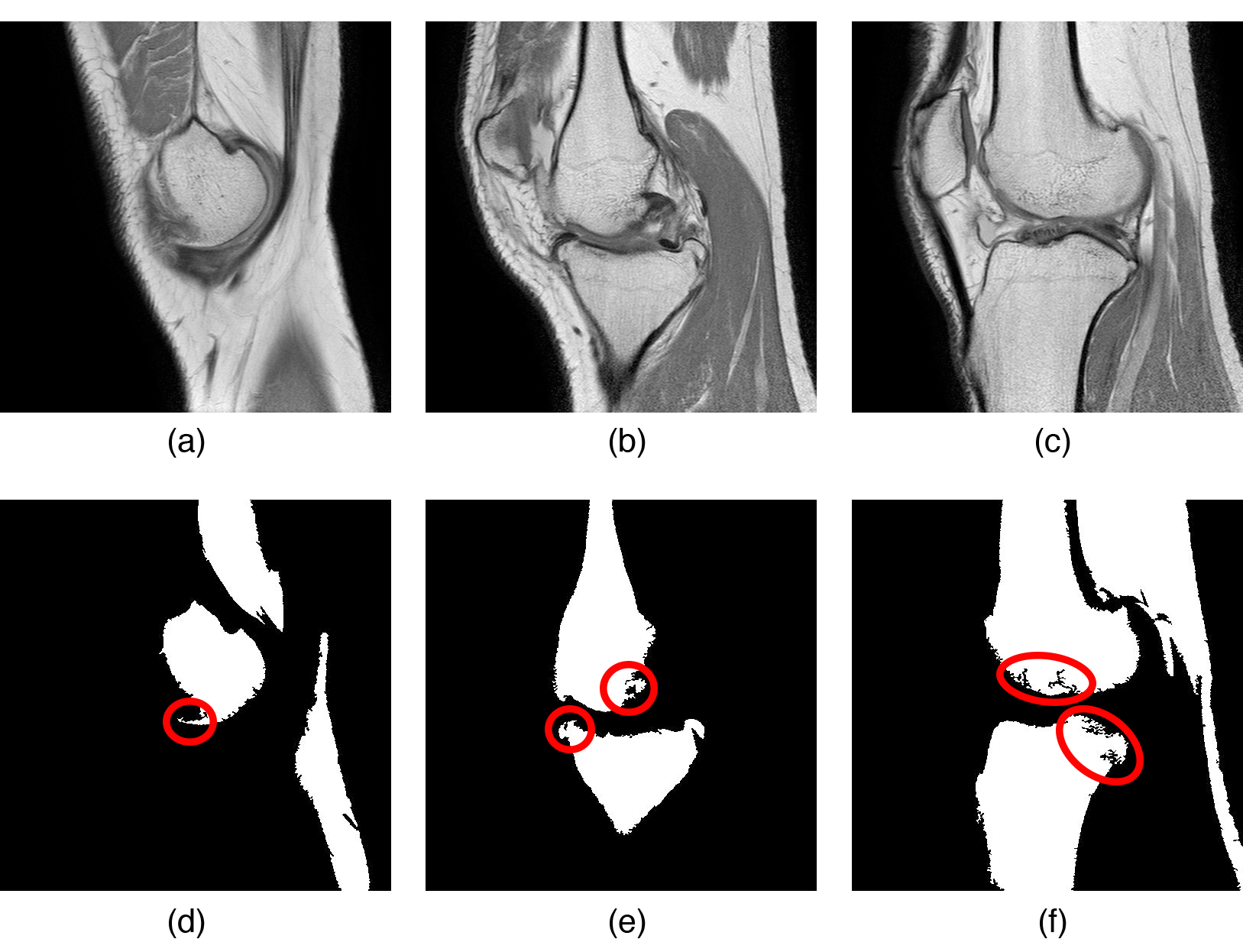}
  \caption{(a), (b) \& (c): Different input images; (d), (e) \& (f): Bottom Row: Initial marker $M$ using regional maxima on above respective input images.}
  \label{bad}
 \end{figure}
However in the result obtained by regional maxima, the initial marker has some of the mostly-occluded objects not marked. Also, the foreground markers in some objects go right up to the object's edge as can be seen in Fig. \ref{bad}. That means we need to clean the edges of the marker blobs and then shrink them a bit. This is done by a closing followed by an erosion operation. Morphological closing of image $M$ by structuring element $B$ denoted by $ M \bullet B $ is defined as:\\
\begin{equation} \label{eq11}
  M \bullet B = (M \oplus B)\ominus B 	
 \end{equation}
 Here, $\bullet$ is the closing operator, $\oplus$ represents dilation and $\ominus$ represents erosion.\\
 Then erosion operation is applied to the resultant image of Eq. \ref{eq11}. This procedure tends to leave some stray isolated pixels that must be removed. To remove them, we eliminate all connected component having less than 20 pixels. Thus the modified marker image obtained is $M_1$ which will serve as the mask image for the stage two of the framework which uses the GrabCut algorithm.\par

\begin{figure}[h]
\centering
\includegraphics[scale=0.15]{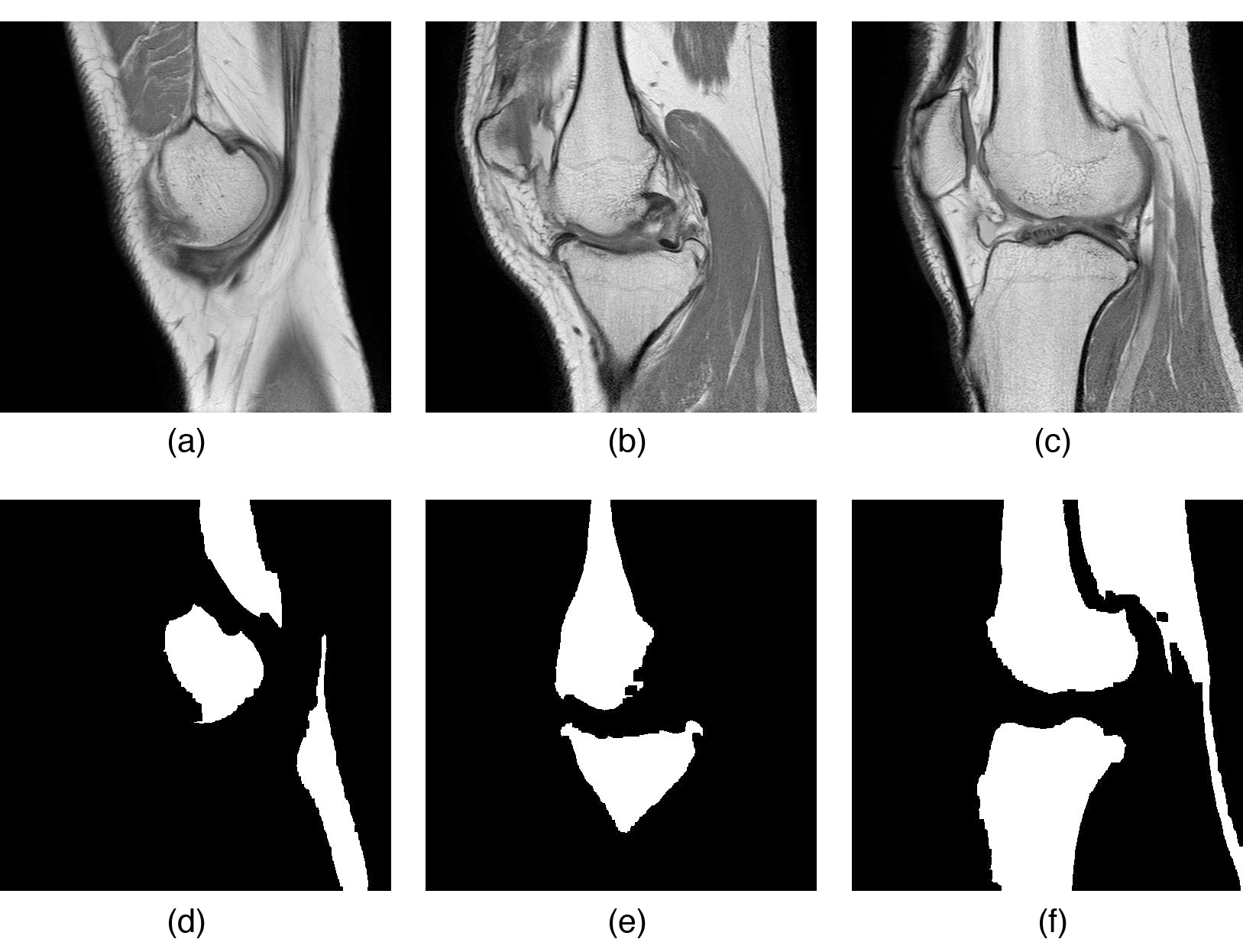}
\caption{(a), (b) \& (c): Different input images; (d), (e) \& (f): Modified marker image $M_1$ on above respective input images.}
\end{figure}
 
\subsection{GrabCut}
GrabCut \cite{ref14} algorithm was designed by Carsten Rother, Vladimir Kolmogorov and Andrew Blake from Microsoft Research Cambridge, UK. It was developed to extract out foreground with minimal user interaction. It is good, both in the segmentation result as well as runtime efficiency.\par 

1) \textit{GrabCut Algorithm}: GrabCut \cite{ref20} is based on the Graph Cut algorithm, which solves the MinCut-MaxFlow problem. An optimization problem is formulated denoting the energy cost function which can be solved by the Graph Cut algorithm. \par
The input to the algorithm are the input image and the labeling. The labeling defines for each pixel whether it belongs to background or foreground. The foreground is obtained by the modified marker and the bounding box aids in providing the background. The aim is to assign lower cost to better labeling. This is done by encouraging same color neighboring pixels to have the same label and vice-versa. Also, the pixels are promoted to match a color distribution model as per their color values. \par
The energy function consists of two parts -- data term and smoothness term. The data term is used to fit the color distribution model. For each pixel, we consider its label $\alpha$ and color $Z$ , and these are provided as input to find out, how well it matches the model using $h()$. The model considered is a K-Gaussian Mixture model. The smoothness term measures the smoothness of the labeling over similar and a-similar neighboring pixels. For every neighboring pair, which does not have similar label, the energy function is increased according to parameter $\beta$ that forms the exponent term and thus determines the smoothness of the labeling. This is captured by a Gibbs energy of the form:

\begin{equation}
E(\underline{\alpha}, \underline{\theta}, z) = U(\underline{\alpha}, \underline{\theta}, z) + V(\underline{\alpha}, z)
\end{equation}

The data term $U$ evaluates the fit of the opacity distribution $\alpha$ to the
data $z$, given the histogram model $\theta$, and is defined as given in Eq. \ref{data}.

\begin{itemize}
\item The data term:
\begin{equation}
U(\underline{\alpha}, \underline{\theta}, z) = \sum_{n} -\log h(z_n;\alpha_n)
\label{data}
\end{equation}
\item The smoothness term:
\begin{equation}
V(\underline{\alpha}, z) = \gamma \sum_{(m,n) \in \textbf{C}} dis(m,n)^{-1} [\alpha_n \neq \alpha_m]   \exp{-\beta(z_m - z_n)^{2}}
\label{smooth}
\end{equation}
\end{itemize}

 As given in Eq. \ref{smooth}, $[\phi]$ denotes the indicator function taking values $0,1$ for a
predicate $\phi$, $C$ is the set of pairs of neighboring pixels and $dis(·)$ is the Euclidean distance of neighbouring pixels. The constant $\beta$ is chosen [Boykov and Jolly 2001] to be:

\begin{equation}
\beta = (2<(z_m - z_n)^{2}>)^{-1}
\end{equation}
where $<.>$ denotes expectation over an image sample. Now that the energy model is fully defined, the segmentation can be estimated as a global minimum:
\begin{equation}
\underline{\hat{\alpha}} =  \arg \min_{\underline{\alpha}} \textbf{E} (\underline{\alpha}, \underline{\theta}). 
\end{equation}

2) \textit{User Interaction}: The output segmented image may have some wrongly classified pixels. To further optimize the result of segmentation, the initial segmented image is refined by the user by marking the wrongly classified foreground as background. Based on the new GMM, the GrabCut algorithm is again run, thereby providing optimized results. Also the user need not mark all the pixels, rather just a small set of pixels in a particular region is enough to provide optimal refined results.

\begin{figure}[h]
\centering
 \includegraphics[scale=0.3]{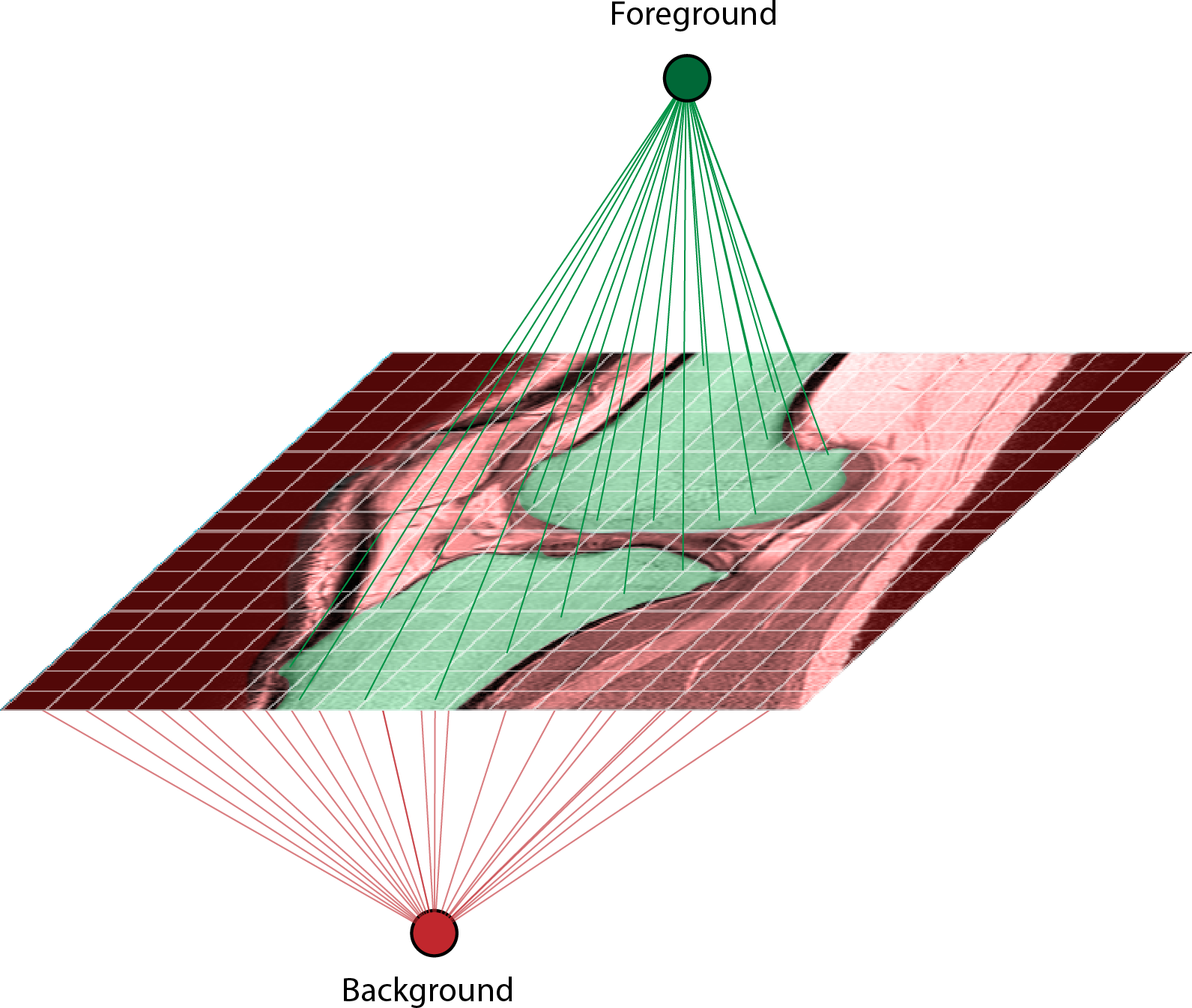}
 \caption{Labeling of pixels as foreground/background for a given MR image.}
\label{grab}
\end{figure}\par

\section{Results and discussion}
\label{section3}
The data set consists mainly of MRI and X-Ray images of different anatomical regions. Also, natural images have been included in the dataset to check the efficiency of MIST in other domains. The MR images in the data set consist of knee images and are provided by All India Institute of Medical Sciences, Jodhpur, India. The format of the MRI dataset is DICOM and the size of each MR image is $512 \times 512$ pixels and the intensity is measured in 16 bits. The X-Ray image dataset is a set of images of different anatomical regions and is a collection of images from the internet. The code is written in MATLAB and C++ using MEX compiler (The implementation is available at \cite{code}. For performance related quantities note that the program is run on a machine having following specifications: Intel Core i5 - 2.7 GHz processor, Windows 8.1, 4 GB RAM. The GUI creation task and the pre-processing work is done in MATLAB R2015a. GrabCut implementation is done using the OpenCV library in C++ and the entire setup in combined using MEX compiler.

\begin{figure}[t]
\centering
\includegraphics[scale=0.15]{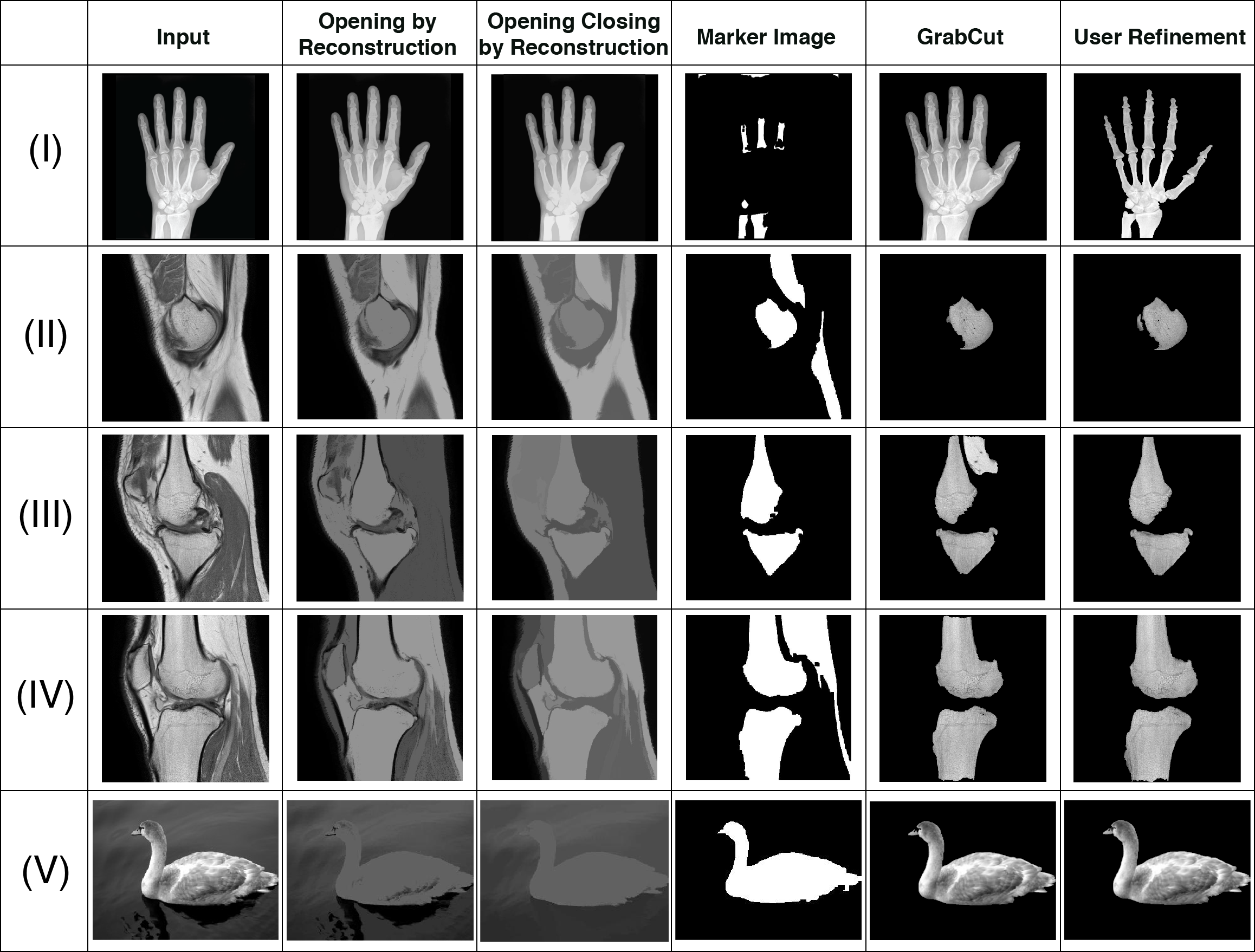}
\caption{Results of intermediate stages of processing in MIST on various images. }
\label{phase}
\end{figure}

The Fig. \ref{phase} shows the output of MIST, along with the intermediate stages. Row (I) refers to a X-ray image; (II), (III) and (IV) refer to MR images; and (V) refers to a natural image. In the subsequent sections, we investigate: (1) the effect of the number of iterations used in the GrabCut algorithm, i.e. for different values of $k$, (2) the effect of the size of the structuring element $B$, (3) comparison with other conventional methods and (4) quantitative evaluation.

\subsection{Effect of the number of iterations in the GrabCut algorithm}

Different values of $k$, i.e. the number of iterations in the GrabCut algorithm, have been experimented to determine the value of $k$ that produces optimal results. It was found that for lower values of $k$, the segmentation results were not accurate. However a large increase in this value increased the running time and did not result into a drastic improvement in the segmentation result as well. Thus a compromise was made in terms of running time and segmentation quality. Thus, based on our experimentation, we chose $k=5$ as an optimal value for efficient segmentation results on the considered dataset. 

\subsection{ Effect of the size of structuring element $B$}

\begin{figure}
\centering
  \includegraphics[width=.24\linewidth]{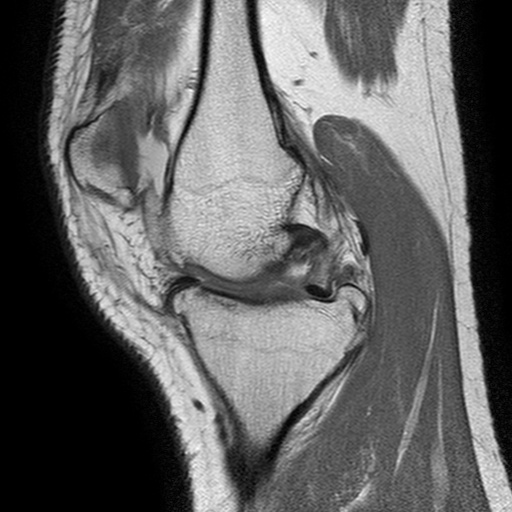}
  \includegraphics[width=.24\linewidth]{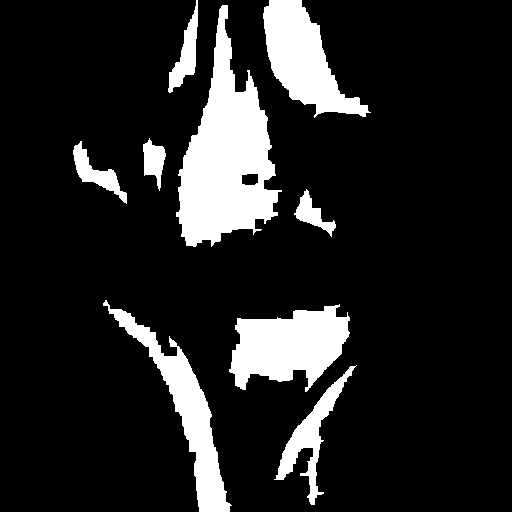}
  \includegraphics[width=.24\linewidth]{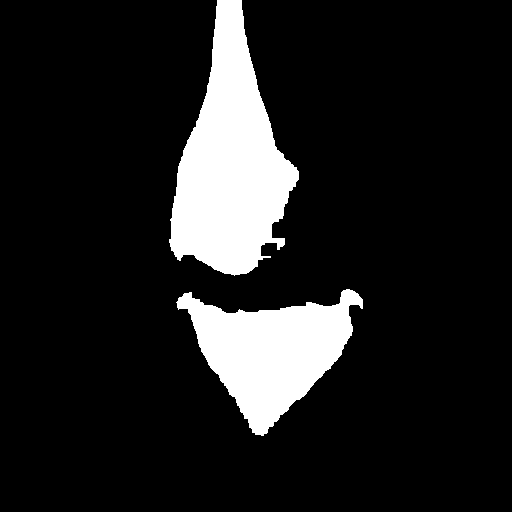}
    \includegraphics[width=.24\linewidth]{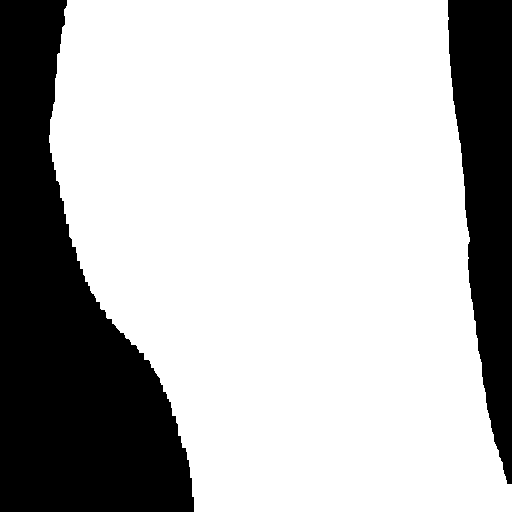}
  \caption{Input image(leftmost); Generated marker images: 10 units, 45 units, 60 units respectively(left to right). }
  \label{markerresult}
\end{figure}

The structuring element $B$ is used to clean up the input image for appropriate foreground marker generation. This structuring element is used for opening-by-reconstruction of the input image. The resultant image is acted upon by closing-by-reconstruction using the same structuring element. Any variation in the size of this structuring element changes the size of the foreground region in the generated marker image. The size of the structuring element also determines the amount of noise removal from the image. For a smaller sized structuring element, even smaller regions are extracted in the marker image. Also, the boundaries in the marker image are not much close to the expected contours to be extracted. While for a structuring element of a comparatively much bigger size, almost the entire region in the image is extracted in the marker image. Thus, for a disk-shaped structuring element of size 45 units i.e. $r=45$ for the considered dataset, the generated marker image has appropriate contours having neither extraction over the boundary region nor much under them. This can be verified by the results shown in Fig. \ref{markerresult}. 

We can see that for $r=10$, many unnecessary smaller regions have been segmented out whereas for $r=60$, segmentation is over the entire image. Thus, a heuristically chosen value of $r=45$ gives an optimal result for the considered dataset.

\subsection{Comparison with other conventional methods}

\begin{figure}[t]
\centering
\includegraphics[scale=0.13]{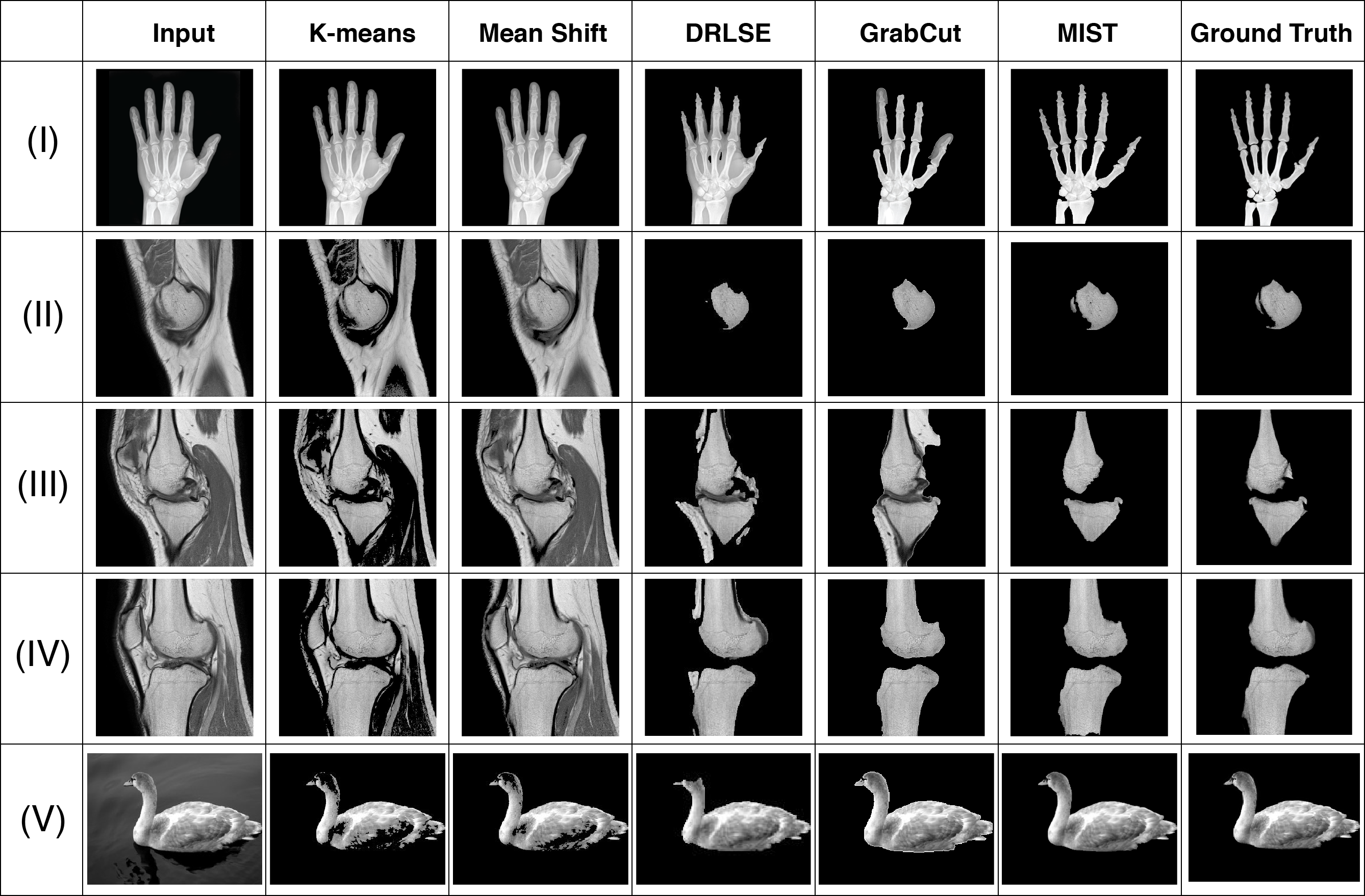}
\caption{Comparison with other conventional methods for different images along with the ground truth. }
\label{compare}
\end{figure}

In this section, the segmentation results of MIST is compared with other conventional techniques. All the results were verified with respect to the ground truth image obtained via hand labelling which was later validated by the doctor. K-means clustering algorithm \cite{ref11} clusters the data by iteratively computing mean intensity for each class and segmenting the image by classifying each pixel in the class with the closest mean. The number of classes were taken as $2$, viz. foreground and background. The obtained two class image was superimposed with the input image for appropriate visualization. We observe in Fig. \ref{compare} that the results obtained by k-means largely vary with respect to the ground truth image. Mean shift clustering \cite{ref12} is a general non-parametric technique used to analyze multimodal feature space. The results from mean shift algorithm are better as compared to k-means clustering algorithm but still substantially different from the ground truth image. Distance Regularized Level Set Evolution (DRLSE) \cite{ref13} is a variational level set formulation in which the regularity of the level set function is intrinsically maintained during the level set evolution. DRLSE allows the use of a more general and efficient initialization of level set function. For more details, level set evolution \cite{ref17} and DRLSE \cite{ref18} can be referred by the readers. Fig. \ref{compare} shows the results of DRLSE which on comparison with k-means clustering algorithm and mean shift clustering algorithm, is better and more accurate but still lag behind the optimal results. The GrabCut algorithm extends the Graph cuts technique proposed in \cite{ref7} through iterative energy minimization process. GrabCut \cite{ref14} uses the MRF formulation of the segmentation problem like the Graph cuts technique and employs min-cut max-flow algorithm. Fig. \ref{compare} shows the results of applying GrabCut algorithm on the input images. The results of MIST are shown in Fig. \ref{compare} and it clearly outperforms the original GrabCut algorithm along with other methods as well. Also the obtained image closely resembles the ground truth image in all the cases. For some complex scenarios, the user requires a bit user interaction but the time consumption for overall segmentation is still less. To get a deeper insight into the accuracy of MIST, qualitative results are provided on two images obtained by different modalities, X-Ray and MRI.

\begin{figure}[t]
\centering
\includegraphics[scale=0.16]{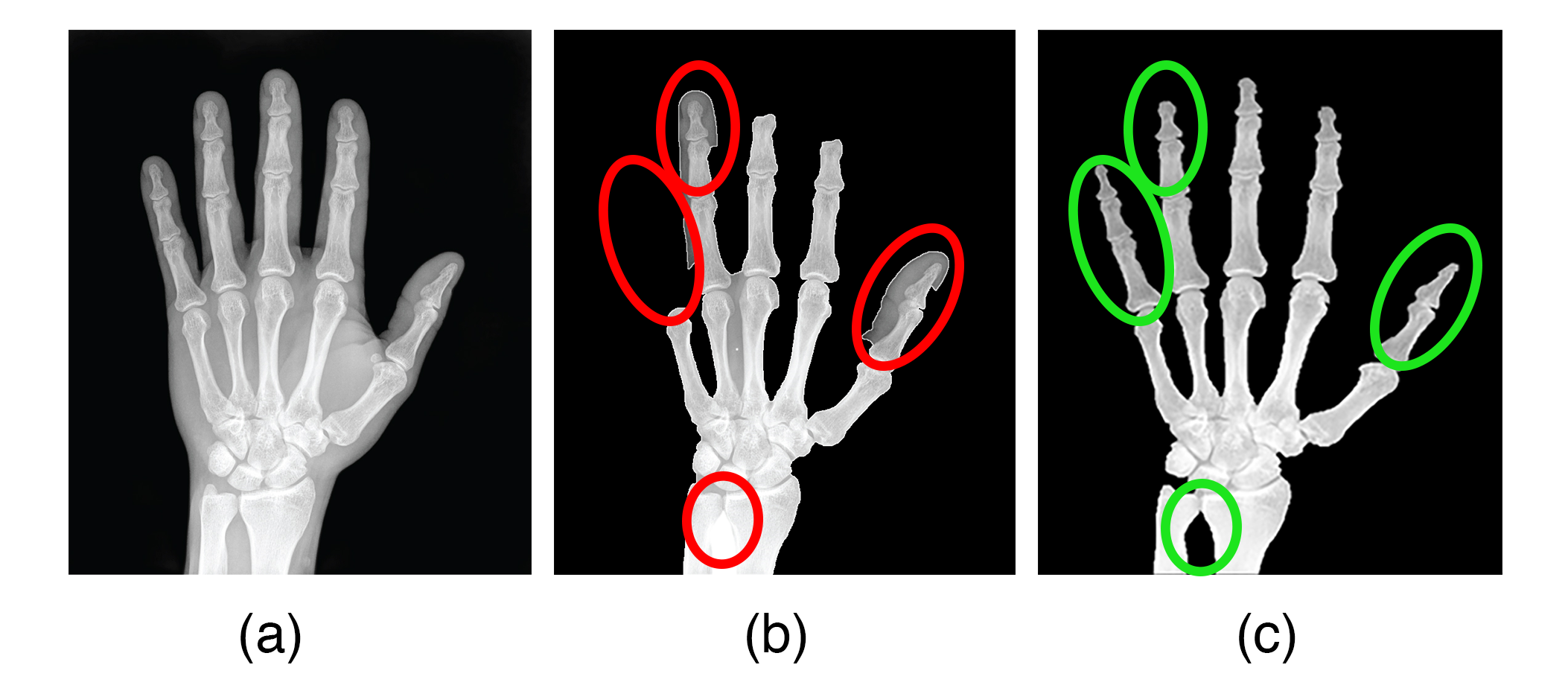}
\caption{(a) Input X-ray image, (b) Only GrabCut \cite{ref14} result, (c) result of MIST. }
\label{corr}
\end{figure}
Fig. \ref{corr}(a) shows a X-ray image of a hand. Fig. \ref{corr}(b)  and Fig. \ref{corr}(c) are segmented outputs by applying only grabcut and that of MIST respectively. We can clearly see that the regions marked in red in Fig. \ref{corr}(b) are improperly segmented. Even the distal phalanx and the middle phalanx of the little finger is also missing in the result obtained by applying only GrabCut algorithm. These minute details are clearly retained in the segmented output of MIST as marked in green color in Fig. \ref{corr}(c). Also, the tip of the fingers are accurately segmented in MIST.

\begin{figure}[t]
\centering
\includegraphics[scale=0.16]{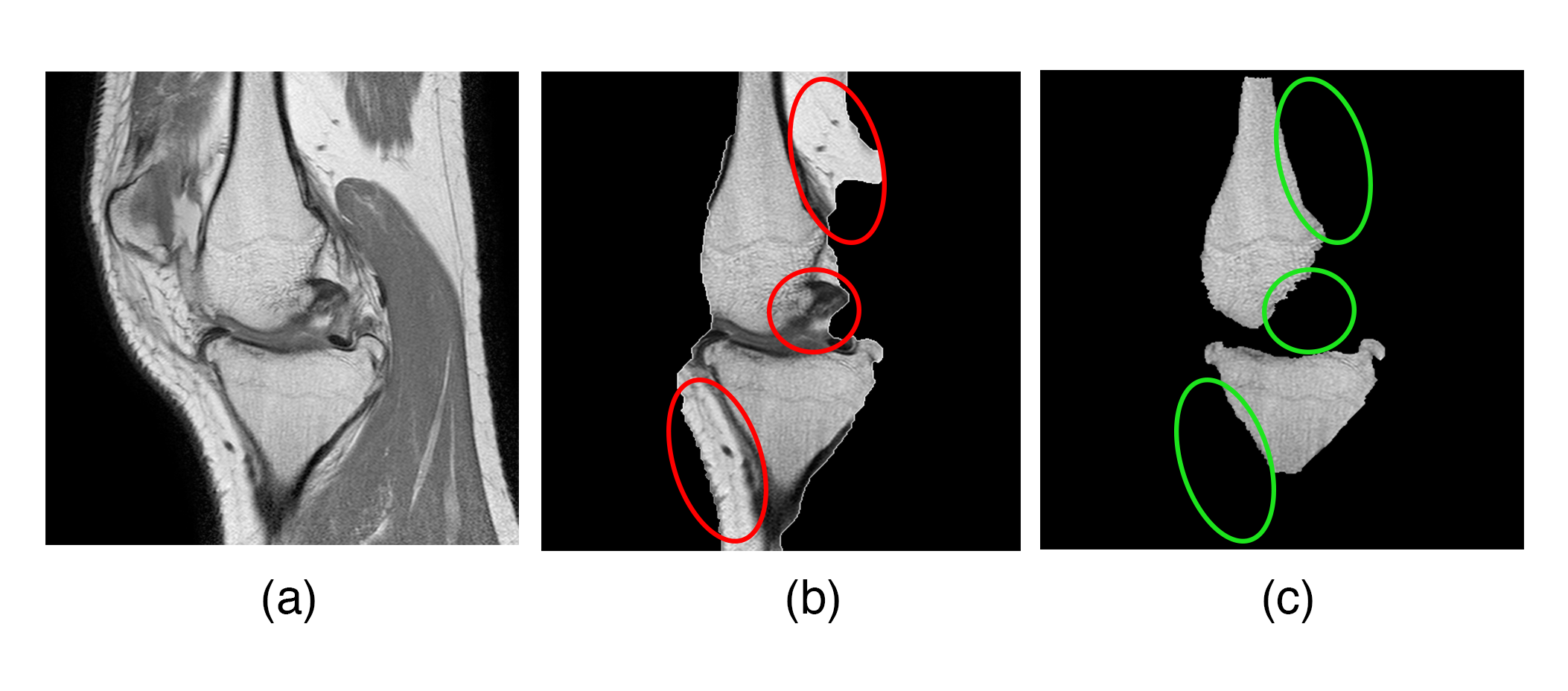}
\caption{(a) Input MR image, (b) Only GrabCut \cite{ref14} result, (c) result of MIST.}
\label{correc}
\end{figure}

Fig. \ref{correc}(a) shows a MR image of a knee joint. We can clearly see in Fig. \ref{correc}(b) that the regions marked in red are the tissue regions and are improperly segmented out. The femur and tibia bones have adjacent muscle regions(marked in red) also segmented out in the results obtained by applying only the GrabCut algorithm. On comparing with MIST as shown in Fig. \ref{correc}(c) we find that MIST yields a better result of segmenting the bone regions more efficiently and accurately without including any kind of tissue. The shortcomings of the GrabCut method are clearly overcome by MIST as it can be seen by the regions marked as green.

\subsection{Quantitative evaluations}
The qualitative analysis clearly depicted the efficacy of MIST in the previous section. Quantitative analysis in terms of accuracy and running time of the proposed algorithm will provide a better analysis of the comparison between the proposed method and other conventional methods \cite{ref11, ref12, ref13, ref14}. The further two subsections provide an insight into the comparison of the methods in terms of accuracy and running time of the proposed algorithm:

\subsubsection{Accuracy}

To evaluate the performance of our segmentation approach, we have used the Dice similarity coefficient index and Hausdorff distance as a measure to quantify the consistency between the segmented results and the ground truth which has been validated by the doctors. The Dice coefficient is given by:

\begin{equation}
d = 2 * \dfrac{|R_{seg} \cap R_{gt}|}{|R_{seg}| + |R_{gt}|}  
\end{equation}

where $R_{seg}$ is the segmented result of MIST and $R_{gt}$ is the manually segmented ground truth.\\ 

On the other hand the Hausdorff distance $d$ between two surfaces $\mathcal{S}_1,\mathcal{S}_2$ is defined as the maximum of the two relative distances:

\begin{equation}
  d(\mathcal{S}_1,\mathcal{S}_2) = \max \left\{ \Delta(\mathcal{S}_1,\mathcal{S}_2), \Delta(\mathcal{S}_2,\mathcal{S}_1)\right\}
\end{equation}

% \usepackage{multirow}
\begin{table}[t]
\label{haus}
\centering
\begin{tabular}{|c|c|c|c|c|c|}
\hline
                                       & \multicolumn{5}{c|}{\textbf{Dice Coefficient}}                                                                           \\ \cline{2-6} 
\multirow{-2}{*}{\textbf{Input Image}} & \multicolumn{1}{c|}{\textbf{\begin{tabular}[c]{@{}c@{}}k-means \\ \cite{ref11} \end{tabular}}}
 & \multicolumn{1}{c|}{\textbf{\begin{tabular}[c]{@{}c@{}}Mean Shift \\ \cite{ref12} \end{tabular}}}
 & \multicolumn{1}{c|}{\textbf{\begin{tabular}[c]{@{}c@{}}DRLSE \\ \cite{ref13} \end{tabular}}}
 & \multicolumn{1}{c|}{\textbf{\begin{tabular}[c]{@{}c@{}}GrabCut \\ \cite{ref14} \end{tabular}}}
               & \multicolumn{1}{c|}{\textbf{\begin{tabular}[c]{@{}c@{}} MIST \end{tabular}}}\\ \hline
   
 \raisebox{-0.5\totalheight}{\includegraphics[scale=0.2]{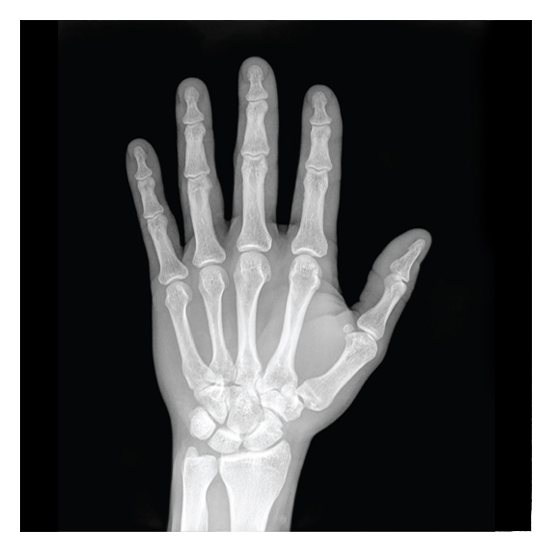}}  & 0.8020           & 0.7948              & 0.8062         & 0.8623                                  & \textbf{0.9074}      \\ \hline
 \raisebox{-0.5\totalheight}{\includegraphics[scale=0.1]{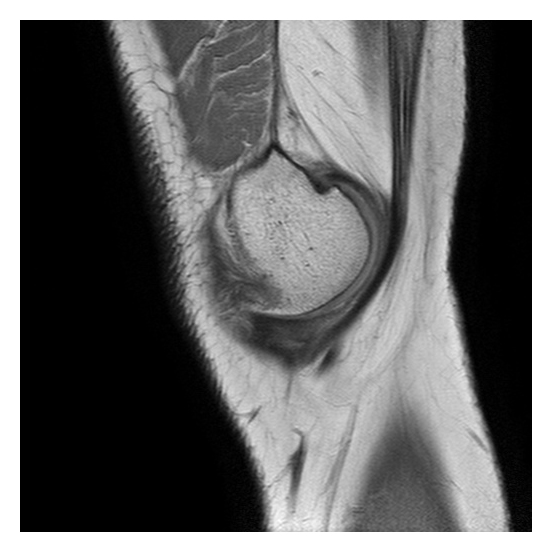}}  & 0.2359           & 0.2106              & 0.8702         & 0.9019                         & \textbf{0.9022}               \\ \hline
\raisebox{-0.5\totalheight}{\includegraphics[scale=0.1]{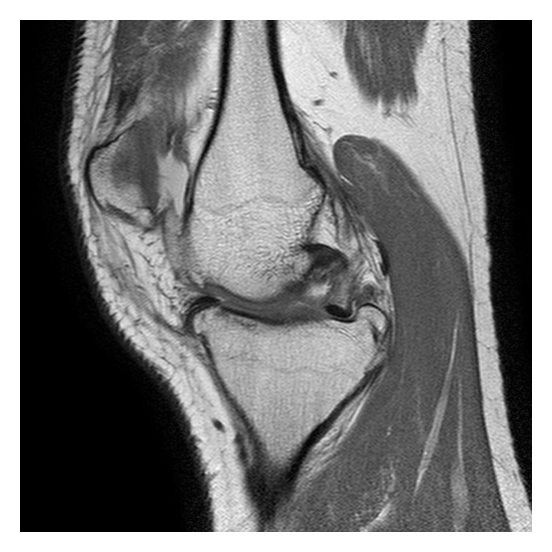}}  & 0.4780           & 0.3854              & 0.7390         & 0.7380                                  & \textbf{0.8961}      \\ \hline
 \raisebox{-0.5\totalheight}{\includegraphics[scale=0.1]{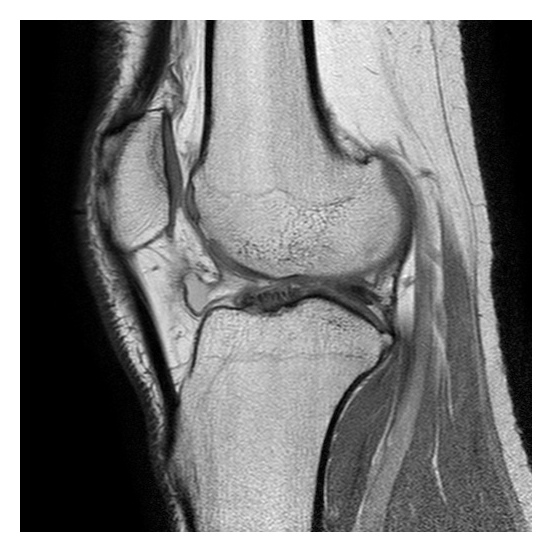}} & 0.6186           & 0.5675              & 0.8933         & 0.9433                                  & \textbf{0.9462}      \\ \hline
\raisebox{-0.5\totalheight}{\includegraphics[scale=0.1]{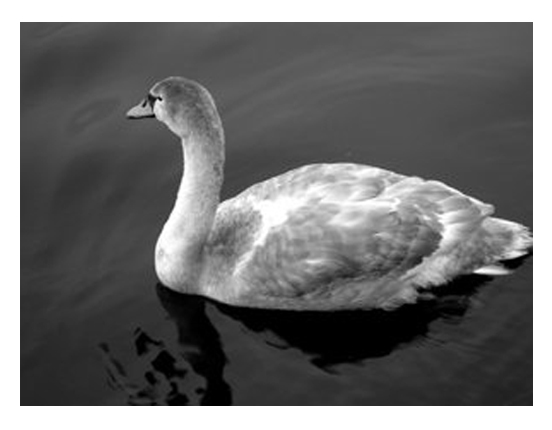}} & 0.8448           & 0.8892              & 0.8905         & \textbf{0.9393} & 0.9389               \\ \hline
\end{tabular}
\caption{Comparison of Dice coefficient for given input images between MIST and other conventional methods. The values in boldface represent the best value among the results of other conventional methods and MIST.}
\label{dice}
\end{table}

% \usepackage{multirow}
\begin{table}[t]
\centering
\begin{tabular}{|c|c|c|c|c|c|}
\hline
 & \multicolumn{5}{c|}{\textbf{Hausdorff Distance}} \\ \cline{2-6} 
\multirow{-2}{*}{\textbf{Input Image}} & \multicolumn{1}{c|}{\textbf{\begin{tabular}[c]{@{}c@{}}k-means \\ \cite{ref11} \end{tabular}}} & \multicolumn{1}{c|}{\textbf{\begin{tabular}[c]{@{}c@{}}Mean Shift \\ \cite{ref12} \end{tabular}}} & \multicolumn{1}{c|}{\textbf{\begin{tabular}[c]{@{}c@{}}DRLSE \\ \cite{ref13} \end{tabular}}} & \multicolumn{1}{c|}{\textbf{\begin{tabular}[c]{@{}c@{}}GrabCut \\ \cite{ref14} \end{tabular}}} & \multicolumn{1}{c|}{\textbf{\begin{tabular}[c]{@{}c@{}} MIST \end{tabular}}}\\ \hline
 \raisebox{-0.5\totalheight}{\includegraphics[scale=0.2]{Image_1_pad.jpg}} & 203.720 & 205.885 & 195.634 & 147.618 & \textbf{83.480} \\ \hline
 \raisebox{-0.5\totalheight}{\includegraphics[scale=0.1]{IM_0005_pad.jpg}} & 202.543 & 243.621 & \textbf{7.071} & 27.023 & 31.0644 \\ \hline
\raisebox{-0.5\totalheight}{\includegraphics[scale=0.1]{IM_0015_pad.jpg}} & 217.444 & 268.306 & 137.790 & 125.726 & \textbf{51.468} \\ \hline
\raisebox{-0.5\totalheight}{\includegraphics[scale=0.1]{IM_0025_pad.jpg}} & 206.838 & 280.592 & 97.00 & 41.857 & \textbf{27.893} \\ \hline
    \raisebox{-0.5\totalheight}{\includegraphics[scale=0.1]{duck_pad.jpg}} & 30.5450 & 32.435 & \textbf{31.733} & 40.062 & 35.440 \\ \hline
\end{tabular}
\caption{Comparison of Hausdorff distance for given input images between MIST and other conventional methods. The values in boldface represent the best value among the results of other conventional methods and MIST. }
\end{table}

Table \ref{dice} and \ref{haus} shows the comparison of Dice similarity coefficient index and Hausdorff distance calculated for the segmented results obtained by the methods for each of the input images shown. The numerical values marked in boldface represent the best value among all the respective values.

\begin{table}[t]
\centering
\begin{tabular}{|l|c|c|}
\hline
\multicolumn{1}{|c|}{\textbf{Methods}} & \textbf{Dice Coefficient} & \textbf{Hausdorff Distance} \\ \hline
k-means clustering \cite{ref11} & 0.5249 & 169.745 \\ \hline
Mean shift clustering \cite{ref12} & 0.3927 & 221.744 \\ \hline
DRLSE \cite{ref13} & 0.7104 & 95.0747 \\ \hline
GrabCut \cite{ref14} & 0.8311 & 42.629 \\ \hline
MIST & \textbf{0.8545} & \textbf{34.375} \\ \hline
\end{tabular}
\caption{Comparison of Dice coefficient and Hausdorff distance between MIST and other conventional methods.}
\label{com}
\end{table}

The above metric was computed for all the images in the dataset and averaged out for different algorithms. As shown in Table \ref{com}, MIST stands out in both the metrics as highlighted by boldface.

\subsubsection{Performance analysis based on running time} 

The running time is one of the major factors necessary to be considered along with accuracy in the field of medical imaging for real time output generation. The time has been calculated using the CPU clock cycle time of the system.

% \usepackage{multirow}

\begin{table}[b]
\centering
\begin{tabular}{|c|c|c|c|c|c|}
\hline
\textbf{Methods} & \multicolumn{1}{c|}{\textbf{\begin{tabular}[c]{@{}c@{}}k-means \\ \cite{ref11} \end{tabular}}} & \multicolumn{1}{c|}{\textbf{\begin{tabular}[c]{@{}c@{}}Mean Shift \\ \cite{ref12} \end{tabular}}} & \multicolumn{1}{c|}{\textbf{\begin{tabular}[c]{@{}c@{}}DRLSE \\ \cite{ref13} \end{tabular}}} & \multicolumn{1}{c|}{\textbf{\begin{tabular}[c]{@{}c@{}}GrabCut \\ \cite{ref14} \end{tabular}}} &\multicolumn{1}{c|}{\textbf{\begin{tabular}[c]{@{}c@{}} MIST \end{tabular}}}\\ \hline
\textbf{Time(in sec)}     &  $1.9$                &  $2.593$                   &  $135$              &   $4.2$              &   $4.066$                                                                                    \\ \hline
\end{tabular}
\caption{Average running time computed on the entire dataset for various methods.}
\label{time}
\end{table}

Fig. \ref{time} shows the average time taken to obtain the segmentation results for the entire dataset using various methods. Although the Hausdorff distance for the second image in Table \ref{haus} is best for DRLSE, its running time is substantially high thus making it infeasible for real-time segmentation. The other results are close to each other, so accuracy of the segmented output gives a proper choice of the most appropriate segmentation method. MIST involves user interaction and thus can be used for real time execution. 

\section{Interactive Image Segmentation Software }
\begin{figure}[t]
\centering
\includegraphics[scale=0.25]{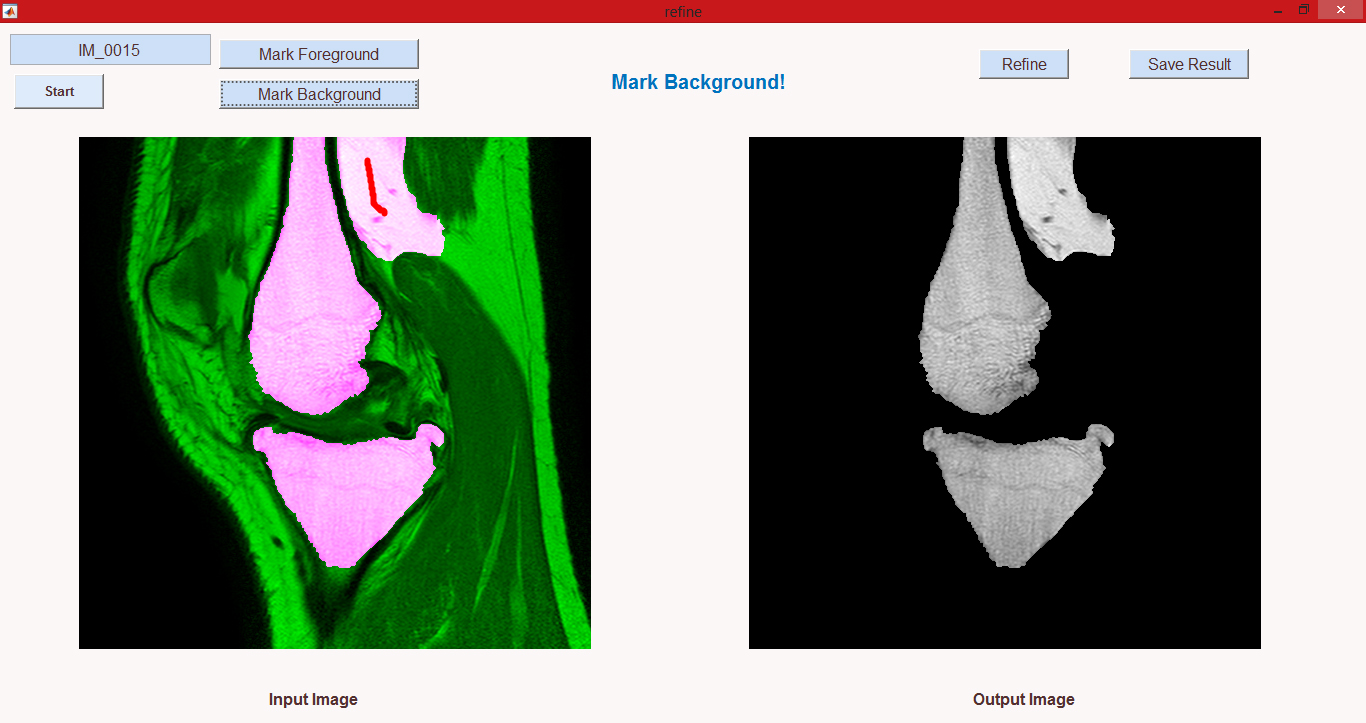}
\caption{MIST incorporated into a GUI.}
\label{final}
\end{figure}

MIST was packaged into a user-friendly Graphical User Interface (GUI) so that the work could be further put to use and provide advancement in the medical field. As mentioned earlier, the GUI is made using MATLAB but the GrabCut code is written in C++ using its OpenCV library. Since the back-end code, i.e. the one involving processing using GrabCut is written in C++, the running time is efficient. The software toolkit can be easily installed and run on any operating system thus making it easily available to the doctors for real-time analysis of the patient data. Fig. \ref{final} shows the overall execution of the program through various stages.
\\
\label{section4}

In Fig. \ref{final}, the input image (left) is first selected by typing the required image name in the top left corner text box. The user can then press the start button, mark a bounding box for the region of interest and the binary marker image is automatically generated in the backend and feeded into the GrabCut to provide the segmented output image on the right. The user can further refine the image (if necessary) by marking the background or foreground region.  This whole procedure takes a span of just a few seconds which may slightly vary from image to image. Since the GUI is simple and provided with instructions on the top, it can be used by any amateur user even for natural images.
\begin{figure}[t]
\centering
\includegraphics[scale=0.15]{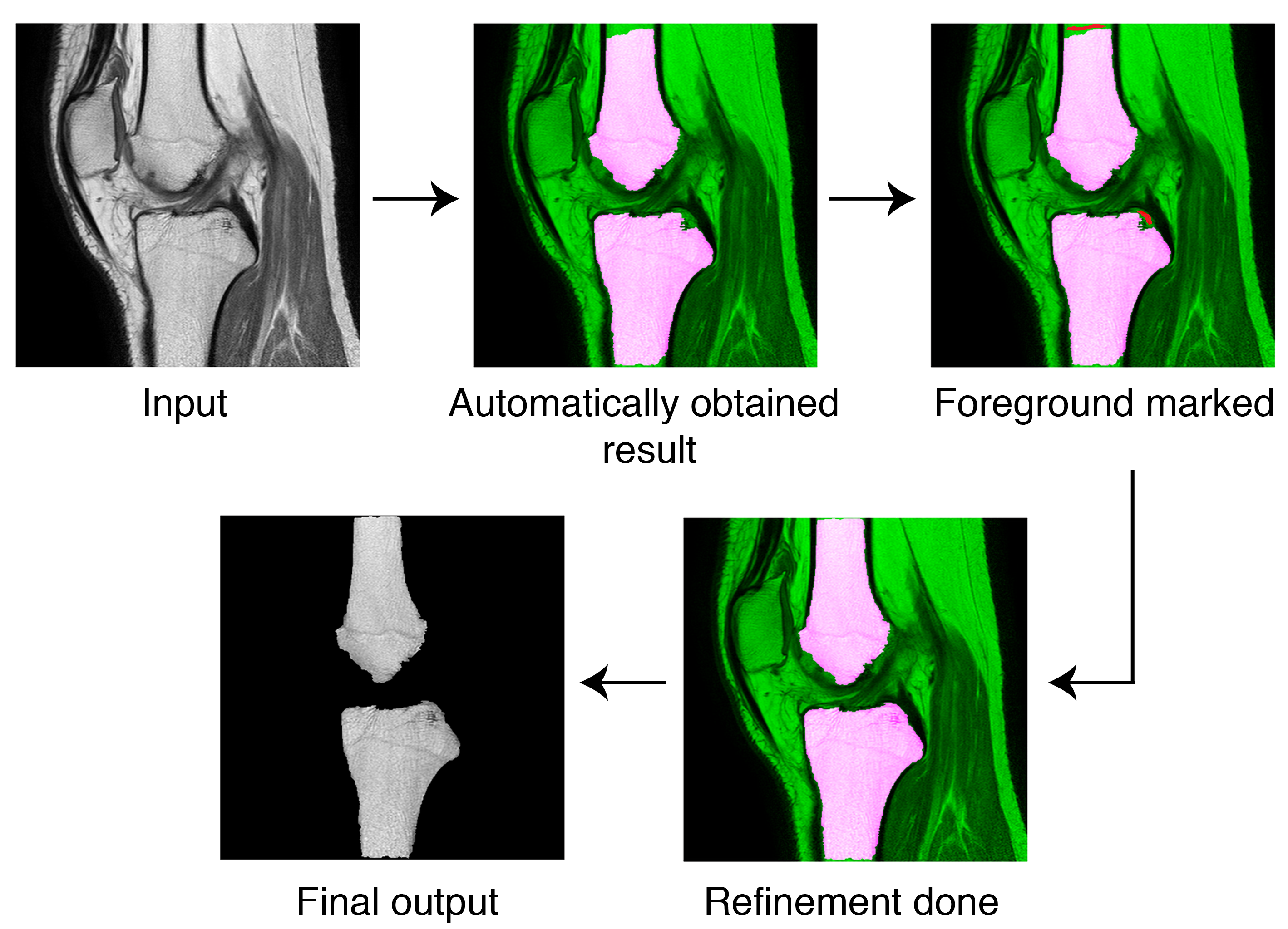}
\caption{Sequential snapshots of results from GUI.}
\label{finalflow}
\end{figure}

Fig. \ref{finalflow} shows the results of sequential steps from our GUI. The green region shows the background whereas the pink region shows the foreground. The input image, result before user interaction, user interaction using red marker and the final output after refinement can be clearly seen respectively in the flow diagram.

\section{Conclusion}
\label{section5}
This paper presents an interactive segmentation technique for medical images based on morphology and GrabCut, combined into a user-friendly GUI using MATLAB.  Initially, the input medical image is pre-processed using morphological operations and then the generated marker image is given as an input to GrabCut algorithm. The generated output is thus subjected to user interaction, if needed, and thus the final segmented output is obtained. The proposed algorithm's results are close to optimal segmentation results and is also time efficient as can be seen in the previous section. The algorithm also works and gives efficient results for natural images and thus can have many applications in various other fields. Thus, MIST provides an excellent alternative not only in the medical field but also in the other fields containing image processing.\par
Future work may involve increasing the accuracy and time efficiency of the algorithm. To increase the accuracy, use of super pixels can be made. Also, in graph creation for GrabCut, instead of choosing each pixel as node, a set of pixels can be chosen as a node. This may increase the computing efficiency.\par

\end{document}